\documentclass[sigconf]{acmart}

\usepackage{booktabs} 
\usepackage[ ]{algorithm2e}
\usepackage[bottom]{footmisc}
\usepackage{balance}

\setcopyright{acmlicensed}
\acmConference[LND4IR '18]{ACM SIGIR 2018 Workshop on Learning from Limited or Noisy Data}{July 12, 2018}{Ann Arbor, Michigan, USA}
\acmYear{2018}
\copyrightyear{2018}
\acmDOI{}
\acmISBN{}
\acmPrice{}

\begin{document}

\title{Multilingual Sentiment Analysis: \\An RNN-Based Framework for Limited Data}

\author{Ethem F. Can}
\affiliation{%
  \institution{SAS Inst.}
}
\email{ethfcan@gmail.com}

\author{Aysu Ezen-Can}
\affiliation{%
  \institution{SAS Inst.}
}
\email{aysu.e.can@gmail.com}

\author{Fazli Can}
\affiliation{%
  \institution{Bilkent IR Group \\ Bilkent University}
}
\email{canf@cs.bilkent.edu.tr}

\begin{abstract}

Sentiment analysis is a widely studied NLP task where the goal is to determine opinions, emotions, and evaluations of users 
towards a product, an entity or a service that they are reviewing. One of the biggest challenges for sentiment analysis is that it is highly language dependent.
Word embeddings, sentiment lexicons, and even annotated data are language specific.
Further, optimizing models for each language is very time consuming and labor intensive 
especially for recurrent neural network models.
From a resource perspective, it is very challenging to collect data for different languages. 

In this paper, we look for an answer to the following research question: \textit{can a sentiment analysis model
trained on a language be reused for sentiment analysis in other languages, Russian, Spanish, Turkish, and Dutch, where the data is more limited?} 
Our goal is to build a single model in the language with the largest dataset available for the task, and reuse it for languages
that have limited resources.
For this purpose, 
we train a sentiment analysis model using recurrent neural networks with reviews in English. 
We then translate reviews in other languages and reuse this model to evaluate the sentiments. 
Experimental results show that our robust approach of single model trained on English reviews statistically significantly 
outperforms the baselines in several different languages.

\end{abstract}

%
%



\maketitle

\section{Introduction}

With the steady growth in the commercial websites and social media venues, the access to users' reviews have become easier. 
As the amount of data that can be mined for opinion increased, commercial companies' interests for sentiment analysis increased as well. 
Sentiment analysis is an important part of understanding user behavior and opinions on products, places, or services.

Sentiment analysis has long been studied by the research community, 
leading to several sentiment-related resources such as sentiment dictionaries that can be used as features for 
machine learning models ~\cite{taboada2011lexicon, banea2008bootstrapping,inui2011applying,steinberger2012creating}. 
These resources help increase sentiment analysis accuracies; however, they are highly dependent on language and require researchers to build such resources for every language to process. 

Feature engineering is a large part of the model building phase for most sentiment analysis and emotion detection models ~\cite{ortigosa2014sentiment}. 
Determining the correct set of features is a task that requires thorough investigation. 
Furthermore, these features are mostly language and dataset dependent making it even further challenging to build models for different languages. 
For example, the sentiment and emotion lexicons, as well as pre-trained word embeddings are not completely transferable to other languages which replicates the efforts 
for every language that users would like to build sentiment classification models on. 
For languages and tasks where the data is limited, extracting these features, building language models, training word embeddings, and creating lexicons are big challenges. 
In addition to the feature engineering effort, the machine learning models' parameters also need to be tuned separately for each language to get the optimal results. 

In this paper, we take a different approach. We build a reusable sentiment analysis model that does not utilize any lexicons. 
Our goal is to evaluate how well a generic model can be used to mine opinion in different languages where data is more limited than the language where the generic model is trained on. 
To that end, we build a training set that contains reviews from different domains in English (e.g., movie reviews, product reviews) and train a recurrent neural network (RNN) 
model to predict polarity of those reviews. Then focusing on a domain, we make the model specialized in that domain by using the trained weights from the larger data and 
further training with data on a specific domain. To evaluate the reusability of the sentiment analysis model, we test with non-English datasets. 
We first translate the test set to English and use the pre-trained model to score polarity in the translated text. 
In this way, our proposed approach eliminates the need to train language-dependent models, use of sentiment lexicons and word embeddings for each language. 
Our experiments show that a generalizable sentiment analysis model can be utilized successfully to perform opinion mining for languages that do not have enough resources to train specific models. 

The contributions of this study are; 1) a robust approach that utilizes machine translation to reuse a model trained on one language in other languages, 
2) an RNN-based approach to eliminate feature extraction as well as resource requirements for sentiment analysis, and 
3) a technique that statistically significantly outperforms baselines for multilingual sentiment analysis task when data is limited. 
To the best of our knowledge, this study is the first to apply a deep learning model to the multilingual sentiment analysis task.

\section{Related Work}

There is a rich body of work in sentiment analysis including social media platforms such as Twitter ~\cite{pak2010twitter} and Facebook ~\cite{ortigosa2014sentiment}. 
One common factor in most of the sentiment analysis work is that features that are specific to sentiment analysis are extracted (e.g., sentiment lexicons) and used in different 
machine learning models. 
Lexical resources ~\cite{taboada2011lexicon, banea2008bootstrapping, ortigosa2014sentiment} for sentiment analysis such as SentiWordNet ~\cite{denecke2008using, ahmad2006multi}, 
linguistic features and expressions ~\cite{boiy2009machine}, polarity dictionaries ~\cite{inui2011applying, steinberger2012creating}, other features 
such as topic-oriented features and syntax ~\cite{pang2008opinion}, emotion tokens ~\cite{cui2011emotion}, word vectors ~\cite{maas2011learning}, and 
emographics ~\cite{volkova2013exploring} are some of the information that are found useful for improving sentiment analysis accuracies. 
Although these features are beneficial, extracting them requires language-dependent data (e.g., a sentiment dictionary for Spanish is trained on Spanish data instead 
of using all data from different languages).

Our goal in this work is to streamline the feature engineering phase by not relying on any dictionary other than English word embeddings that are trained on any 
data (i.e. not necessarily sentiment analysis corpus). To that end, we utilize off-the-shelf machine translation tools to first 
translate corpora to the language where more training data is available and use the translated corpora to do inference on.

Machine translation for multilingual sentiment analysis has also seen attention from researchers. Hiroshi et al. ~\cite{hiroshi2004deepera} translated only sentiment units with a pattern-based approach.
Balahur and Turchi ~\cite{balahur2014comparative} used uni-grams, 
bi-grams and tf-idf features for building support vector machines on translated text. Boyd-Graber and Resnik ~\cite{BoydGraber} built Latent Dirichlet 
Allocation models to investigate how multilingual concepts are clustered into topics. Mohammed et al. ~\cite{Mohammad:2016:TAS:3013558.3013562} translate 
Twitter posts to English as well as the English sentiment lexicons. Tellez et al. ~\cite{tellez2017simple} propose a framework where language-dependent 
and independent features are used with an SVM classifier. These machine learning approaches also require a feature extraction phase where 
we eliminate by incorporating a deep learning approach that does the feature learning intrinsically. 
Further, Wan~\cite{wan2008using} uses an ensemble approach
where the resources (e.g., lexicons) in both the original language and the translated language
are used -- requiring resources to be present in both languages. Brooke et al.~\cite{brooke2009cross} also
use multiple dictionaries. 

In this paper, we address the resource bottleneck of 
these translation-based approaches and propose a deep learning approach
that does not require any dictionaries.

\section{Methodology}

\begin{figure}
\centering
\fbox{\includegraphics[width=0.46\textwidth]{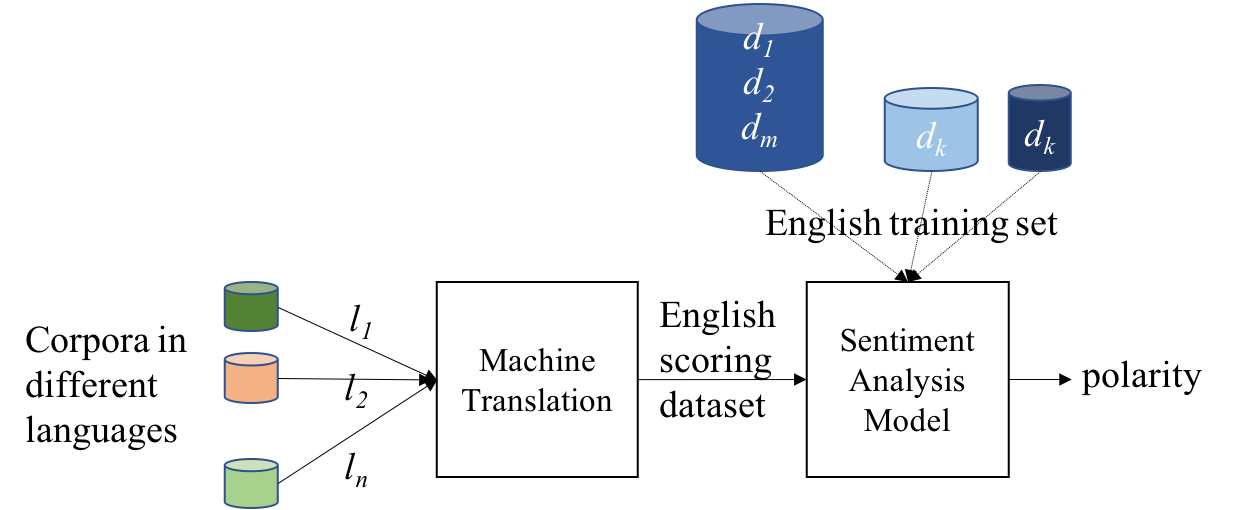}}
\caption{Multilingual sentiment analysis approach.
  }
 \label{fig:framework}
\end{figure}

In order to eliminate the need to find data and build separate models for each language, we propose a multilingual approach where a single model is built in the language where the largest resources are available. In this paper we focus on English as there are several sentiment analysis datasets in English. To make the English sentiment analysis model as generalizable as possible, we first start by training with a large dataset that has product reviews for different categories. Then, using the trained weights from the larger generic dataset, we make the model more specialized for a specific domain. We further train the model with domain-specific English reviews and use this trained model to score reviews that share the same domain from different languages. To be able to employ the trained model, test sets are first translated to English via machine translation and then inference takes place. Figure \ref{fig:framework} shows our multilingual sentiment analysis approach. 
It is important to note that this approach does not utilize any resource in any of the languages of the test sets (e.g., word embeddings, lexicons, training set).

Deep learning approaches have been successful in many applications ranging from computer vision to natural language processing \cite{alom2018history}. 
Recurrent neural network (RNN) including Long Short Term Memory (LSTM) and Gated Recurrent Units (GRU) are subsets of deep learning algorithms where the 
dependencies between tokens can be used by the model. These models can also be used with variable length input vectors which makes them suitable for text input. 
LSTM and GRU models allow operations of sequences of vectors over time and have the capability to `remember' previous information \cite{alom2018history}. 
RNN have been found useful for several natural language processing tasks including language modeling, text classification, machine translation. 
RNN can also utilize pre-trained word embeddings (numeric vector representations of words trained on unlabeled data) without requiring hand-crafted features. 
Therefore in this paper, we employ an RNN architecture that takes text and pre-trained word embeddings as inputs and generates a classification result. 
Word embeddings represent words as numeric vectors and capture semantic information. They are trained in an unsupervised fashion making it useful for our task. 

\begin{figure}
\centering
\fbox{\includegraphics[width=0.46\textwidth]{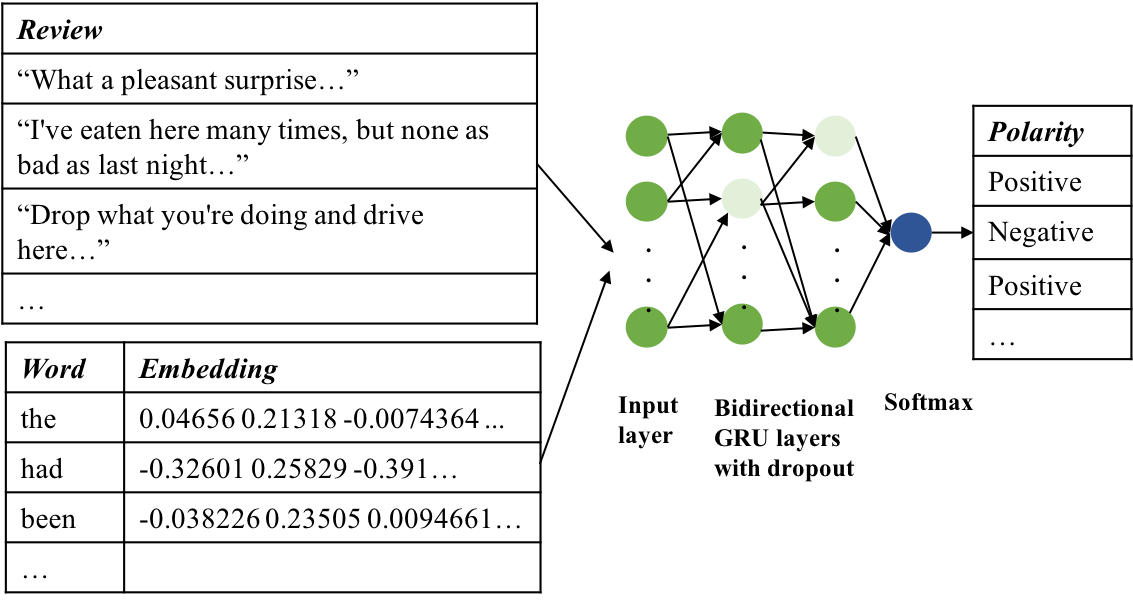}}
\caption{Training sentiment analysis model with RNN.
  }
 \label{fig:trainingPhase}
\end{figure}

The sentiment analysis model that is trained on English reviews has two bidirectional layers, each with 40 neurons and a dropout ~\cite{srivastava2014dropout} of 0.2 is used. 
The training phase takes pre-trained word embeddings and reviews in textual format, then predicts the polarity of the reviews. 
For this study, an embedding length of 100 is used (i.e., each word is represented by a vector of length 100). We utilized pre-trained global vectors \cite{pennington2014glove}.
The training phase is depicted in Figure~\ref{fig:trainingPhase}.

\section{Experiments}
To evaluate the proposed approach for multilingual sentiment analysis task, we conducted experiments. This section first presents the corpora used in this study followed by experimental results. 

Throughout our experiments, we use SAS Deep Learning Toolkit. For machine translation, Google translation API is used.

\subsection{Corpora}
Two sets of corpora are used in this study, both are publicly available. The first set consists of English reviews and the second set contains 
restaurant reviews from four different languages (Spanish, Turkish, Dutch, Russian). We focus on polarity detection in reviews, therefore all datasets in 
this study have two class values (positive, negative). 

\subsubsection{Training Sets}

With the goal of building a generalizable sentiment analysis model, we used three different training sets as provided in Table ~\ref{training_data}. 
One of these three datasets (Amazon reviews ~\cite{he2016, amazonReviewsWebsite}) is larger and has product reviews from several different categories 
including book reviews, electronics products reviews, and application reviews. The other two datasets are to make the model more specialized in the domain. 
In this paper we focus on restaurant reviews as our domain and use Yelp restaurant reviews dataset extracted from Yelp Dataset Challenge ~\cite{yelpWebsite} and restaurant reviews 
dataset as part of a Kaggle competition ~\cite{kaggleWebsite}.

\begin{center}
\begin{table}[t]
\begin{tabular}{ c r } 
\hline
Dataset name &  \# of observations \\
 \hline
Amazon reviews   &  $9,478,095$ \\ 
Yelp restaurant reviews & $8,539$  \\
Competition restaurant reviews  & $68,170$  \\  
\hline
\end{tabular}
\caption{\label{training_data} Datasets used for training. }
\end{table}
\end{center}

\subsubsection{Test Sets}

For evaluation of the multilingual approach, we use four languages. These datasets are part of SemEval-2016 Challenge Task 5 ~\cite{pontiki2016semeval,semevalWebsite}. Table ~\ref{test_data} shows the number of observations in each test corpus.

\begin{table}
\begin{tabular}{ c c r  } 
\hline
Dataset name &  Description & \# of observations \\
 \hline 
$s\_r$ & Spanish restaurant reviews  &  $2,045$ \\
$t\_r$ & Turkish restaurant reviews & $932$ \\ 
$d\_r$ & Dutch restaurant reviews & $1,635$  \\ 
$r\_r$ & Russian restaurant reviews & $2,529$  \\ 
\hline
\end{tabular}
\caption{\label{test_data} Datasets used for testing. }
\end{table}

\subsection{Experimental Results}
For experimental results, we report majority baseline for each language where the majority baseline corresponds to a model's accuracy if it always predicts the majority class in the dataset. For example, if the dataset has 60\% of all reviews positive and 40\% negative, majority baseline would be 60\% because a model that always predicts ``positive'' will be 60\% accurate and will make mistakes 40\% of the time.

In addition to the majority baseline, we also compare our results with a lexicon-based approach. 
We use SentiWordNet~\cite{baccianella2010sentiwordnet} to obtain a positive and a negative sentiment score for each token in a review. 
Then sum of positive sentiment scores and negative sentiment scores for each review is obtained by summing up the scores for each token. 
If the positive sum score for a given review is greater than the negative sum score, we accept that review as a positive review. 
If negative sum is larger than or equal to the positive sum, the review is labeled as a negative review. 

RNN outperforms both baselines in all four datasets (see Table ~\ref{experimental_results}). 
Also for Spanish restaurant review, the lexicon-based baseline is below the majority baseline which shows that solely translating data and using lexicons is not sufficient to 
achieve good results in multilingual sentiment analysis.

\begin{table}
\begin{tabular}{ c c c c  } 
\hline
Dataset & Majority Baseline & Lexicon-based Baseline & RNN  \\
 \hline 
$s\_r$ & 72.71 & 70.98 &  \textbf{84.21} \\
$t\_r$ &  56.97 & 61.59 &  \textbf{74.36} \\ 
$d\_r$ & 59.63 & 70.52  & \textbf{81.77}   \\ 
$r\_r$ & 79.60 & 67.81  & \textbf{85.61}   \\ 
\hline
\end{tabular}
\caption{\label{experimental_results} Accuracy results (\%) for RNN-based approach compared with majority baseline and lexicon-based baseline. }
\end{table}

Among the wrong classifications for each test set, we calculated the percentage of false positives and false negatives. Table ~\ref{experimental_results2} shows the distribution of false positives and false negatives for each class. In all four classes, the number of false negatives are more than the number of false positives. This can be explained by the unbalanced training dataset where the number of positive reviews are more than the number of negative reviews (59,577 vs 17,132).

\begin{table}
\begin{tabular}{ c c c c  } 
\hline
Dataset & False positives & False negatives  \\
 \hline 
$s\_r$ & 30.03 & 69.97   \\
$t\_r$ & 18.83 & 81.17  \\ 
$d\_r$ &  13.42 & 86.58  \\ 
$r\_r$ &  35.16 & 64.84  \\ 
\hline
\end{tabular}
\caption{\label{experimental_results2} Percentage of false positives and false negatives of wrong classifications. }
\end{table}

To be able to see the difference between baseline and RNN, we took each method's results as a  group (4 values: one for each language) and compared the means. 
Post hoc comparisons using the Tukey HSD test indicated that the mean accuracies for baselines (majority and lexicon-based) are significantly different than RNN accuracies as can 
be seen in Table ~\ref{anova_table} (family-wise error rate=0.06). When RNN is compared with lexicon-based baseline and majority baseline, 
the null hypothesis can be rejected meaning that each test is significant. In addition to these comparisons, we also calculated the effect sizes (using Cohen's d) between the baselines and
our method. The results are aligning with Tukey HSD results such that while our method versus baselines have very large effect sizes, 
lexicon-based baseline and majority baseline have negligible effect size.

\begin{figure}
\centering
\includegraphics[width=0.46\textwidth]{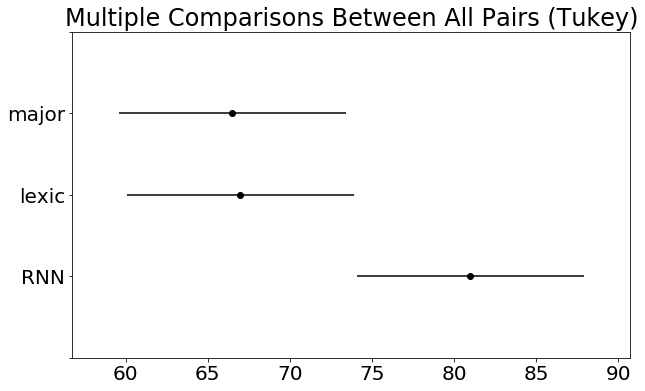}
\caption{Multiple comparisons between majority baseline, lexicon-based baseline and RNN.
  }
 \label{fig:anovaPlot}
\end{figure}

Figure ~\ref{fig:anovaPlot} shows the differences in minimum and maximum values of all three approaches. 
As the figure shows, RNN significantly outperforms both baselines for the sentiment classification task.

\begin{table}
\begin{tabular}{ c r r r r c } 
\hline
group1 & group2 &meandiff&  lower &   upper&  reject\\
\hline
lexicon &  majority  & -0.5  & -14.3168 & 13.3168 & False   \\
lexicon &  RNN    & 14.0&    0.1832 & 27.8168 & True  \\
majority &  RNN &    14.5  &  0.6832 & 28.3168 & True \\
\hline
\end{tabular}
\caption{\label{anova_table} Multiple comparison of means.}
\end{table}

\section{Discussion}
One of the crucial elements while using machine translation is to have highly accurate translations. 
It is likely that non-English words would not have word embeddings, which will dramatically affect the effectiveness of the system.
We analyzed the effect of incorrect translations into our approach. To that end, 
we extracted all wrong predictions from the test set and computed the ratio of misclassifications that have non-English words in them. We first extracted all misclassifications for a given language and for each observation in the misclassification set, we iterated through each token to check if the token is in English. In this way, we counted the number of observations that contained at least one non-English word and divided that with the size of the misclassifications set. We used this ratio to investigate the effect of machine translation errors.


We found that 25.84\% of Dutch,  21.76\% of Turkish, 24.46\% Spanish, and 10.71\% of Russian reviews 
that were misclassified had non-English words in them. These non-English words might be causing the misclassifications. However, a large portion of the 
missclassifications is not caused due to not-translated words. At the end, the machine translation errors has some but not noticeable effects on our model.
Therefore, we can claim that machine translation preserves most of the information necessary for sentiment analysis.

We also evaluated our model with an English corpus~\cite{pontiki2016semeval} to see its performance without any interference from machine translation errors.
Using the English data for testing, the model achieved 87.06\% accuracy where a majority baseline was 68.37\% and the lexicon-based baseline was 60.10\%. 

Considering the improvements over the majority baseline achieved by the RNN model for both non-English (on the average 22.76\% relative improvement; 
15.82\% relative improvement on Spanish, 72.71\% vs. 84.21\%, 30.53\% relative improvement on Turkish, 56.97\% vs. 74.36\%, 37.13\% relative improvement on Dutch, 59.63\% vs. 81.77\%, and
7.55\% relative improvement on Russian, 79.60\% vs. 85.62\%) and English test sets (27.34\% relative improvement), 
we can draw the conclusion that our model is robust to handle multiple languages. 
Building separate models for each language requires both labeled and unlabeled data. 
Even though having lots of labeled data in every language is the perfect case, it is unrealistic. Therefore, eliminating the resource requirement 
in this resource-constrained task is crucial. The fact that machine translation can be used in reusing models from different languages is promising for reducing the data requirements.

\section{Conclusion}
Building effective machine learning models for text requires data and different resources such as pre-trained word embeddings and reusable lexicons. 
Unfortunately, most of these resources are not entirely transferable to different domains, tasks or languages. 
Sentiment analysis is one such task that requires additional effort to transfer knowledge between languages.

In this paper, we studied the research question: \textit{Can we build reusable sentiment analysis models that can be utilized for making inferences in different languages without requiring separate models and resources for each language?} To that end, we built a recurrent neural network model in the language that had largest data available. We took a general-to-specific model building strategy where the larger corpus that had reviews from different domains was first used to train the RNN model and a smaller single-domain corpus of sentiment reviews was used to specialize the model on the given domain. During scoring time, we used corpora for the given domain in different languages and translated them to English to be able to classify sentiments with the trained model. 
Experimental results showed that the proposed multilingual approach outperforms both the majority baseline and the lexicon-based baseline. 

In this paper we made the sentiment analysis model specific to a single domain. 
For future work, we would like to investigate the effectiveness of our model on different review domains including hotel reviews and on different problems such as detecting stance.

\balance

\bibliographystyle{ACM-Reference-Format}
\bibliography{sample-bibliography_multilingual_sentimentAnalysis}

\end{document}